\documentclass[sigconf]{acmart}

\renewcommand\footnotetextcopyrightpermission[1]{}

\AtBeginDocument{%
  }

\setcopyright{acmlicensed}
\copyrightyear{2026}
\acmYear{2026}
\acmDOI{XXXXXXX.XXXXXXX}

\acmConference[KDD 2026]{Proceedings of the 32nd ACM SIGKDD Conference on Knowledge Discovery and Data Mining}{August 03--07, 2026}{Philadelphia, PA, USA}

\acmISBN{978-1-4503-XXXX-X/2026/08}

\usepackage{algorithm}
\usepackage{algorithmic}
\usepackage{graphicx}
\usepackage{textcomp}
\usepackage{xcolor}
\usepackage{booktabs}
\usepackage{multirow}
\usepackage{url}
\usepackage{hyperref}

\usepackage{tabularx}

\begin{document}

\title{SmartMixed: A Two-Phase Training Strategy for Adaptive Activation Function Learning in Neural Networks}


\author{Amin Omidvar}
\affiliation{%
  \institution{Independent Researcher }
  \city{Toronto}
   \state{Ontario}
  \country{Canada}
}
\email{amin@omidvar.cv}

\renewcommand{\shortauthors}{Omidvar}

\begin{abstract}
The choice of activation function plays a critical role in neural networks, yet most architectures still rely on fixed, uniform activation functions across all neurons. We introduce SmartMixed, a novel two-phase training strategy that allows networks to learn optimal per-neuron activation functions while preserving computational efficiency at inference. In the first phase, neurons adaptively select from a pool of candidate activation functions (ReLU, Sigmoid, Tanh, Leaky\_ReLU, ELU, SELU) using a differentiable hard mixture mechanism. In the second phase, each neuron's activation function is fixed according to the learned selection, resulting in a computationally efficient network that supports continued training with optimized vectorized operations. We evaluate SmartMixed on the MNIST dataset using feedforward neural networks of different architectures. Our analysis reveals that neurons in different layers exhibit distinct preferences for activation functions, providing insights into the functional diversity within neural architectures. We also demonstrated that SmartMixed effectively trains the network by allowing neurons to select their preferred activation functions, competing against models using a single fixed state-of-the-art activation function. Code is available online  (\href{https://github.com/omidvaramin/SmartMixed}{link}).
\end{abstract}




\maketitle
\thispagestyle{empty} 
\pagestyle{empty}     

\section{Introduction}
\label{sec:introduction}

The selection of an activation function is an important hyperparameter in designing neural network architectures. Standard practice typically involves applying a single activation function uniformly across all neurons in a model, except for the last layer. Although this approach simplifies implementation, it imposes a rigid structural constraint that prevents individual neurons from adapting their functional behavior to their specific roles within the network.

To address this limitation, recent methods have introduced ensembles of activation functions, such as weighted averages, to provide greater adaptability. Although these approaches allow neurons to express preferences over multiple activation functions, they often introduce substantial computational overhead, such as additional learnable parameters and the need to compute multiple activation functions per neuron rather than just one.

In this paper, we propose \textbf{SmartMixed}, a novel two-phase training strategy that enables efficient per-neuron activation function learning. In the first phase, neurons utilize a differentiable hard-mixture mechanism based on the Gumbel–Softmax estimator to explore a pool of candidate functions, including ReLU, Sigmoid, Tanh, Leaky\_ReLU, ELU, and SELU. In the second phase, the network commits to the most effective learned function for each individual neuron, resulting in an architecture in which each neuron adopts its preferred activation function and is trained using optimized vectorized operations.

The primary contributions of this work are as follows:
\begin{itemize}
    \item We propose a two-phase approach that enables networks to learn optimal per-neuron activation functions while maintaining high computational efficiency.
    \item Our analysis reveals that neurons’ activation function preferences are highly dependent on the network layer, where early layers favor ReLU and Leaky\_ReLU, while deeper layers prefer SELU and ELU.
\end{itemize}

\section{Related Work}
\label{sec:related}

The problem of selecting optimal activation functions has been explored through several approaches. Traditional methods typically assign a single activation function globally across the entire network, with ReLU~\cite{glorot2011deep} and its variants like Leaky\_ReLU~\cite{maas2013rectifier}, ELU~\cite{clevert2015fast}, and SELU~\cite{klambauer2017self} being widely adopted choices.

Several approaches have explored trainable activation functions with learnable parameters~\cite{apicella2021survey}, including parametric ReLU variants and sigmoid-weighted linear units~\cite{elfwing2018sigmoid}. However, these methods typically learn global parameters that affect all neurons similarly. Ramachandran et al.~\cite{ramachandran2017searching} used neural architecture search to discover new activation functions, but applied them uniformly across networks. 

Mixture-based approaches have attempted to combine multiple activation functions. Some works use weighted combinations of activation functions~\cite{dugas2000incorporating} or dynamic selection mechanisms~\cite{manessi2020dynamic}. However, these approaches introduce significant computational overhead by requiring evaluation of multiple activation functions per neuron and additional learnable parameters for mixing weights.

Neural architecture search methods~\cite{liu2018darts} have explored activation function selection as part of broader architectural optimization, but typically focus on layer-wise or block-wise choices rather than individual neuron preferences. Adaptive neural networks~\cite{bengio2013representation} have investigated dynamic architectures, but primarily focus on structural changes rather than activation function adaptation.

Unlike previous approaches, SmartMixed enables each individual neuron to learn and commit to its own optimal activation function from a predefined pool. After the initial learning phase, the resulting network operates with computational efficiency, while maintaining the benefits of per-neuron activation specialization. This fine-grained, neuron-level activation adaptation distinguishes our work from existing methods that operate at network or layer granularity.

\section{Methodology}
\label{sec:methodology}

\subsection{Problem Formulation}

Consider a neural network with layers $l \in \{1, 2, \ldots, L\}$, where layer $l$ contains $n_l$ neurons. In conventional neural networks, all neurons typically share a single activation function $\sigma$, chosen as a global hyperparameter (e.g., ReLU, Sigmoid, or Tanh), with the output layer often being an exception. In contrast, our approach aims to learn an individualized activation function $\sigma_{l,i}$ for each neuron $i$ in layer $l$, selected from a predefined pool $\mathcal{A} = \{\text{ReLU}, \text{Sigmoid}, \text{Tanh}, \text{Leaky\_ReLU}, \text{ELU}, \text{SELU}\}$.

\subsection{Phase 1: Selective Activation Learning}

In the selective phase, each neuron learns to choose its optimal activation function from a pool $\mathcal{A}$ using a differentiable discrete sampling mechanism. For neuron $i$ in layer $l$, we maintain a learnable logit vector $\boldsymbol{\alpha}_{l,i} \in \mathbb{R}^{|\mathcal{A}|}$ that encodes its preference over the candidate activation functions.

To enable stochastic selection, we employ the hard Gumbel--Softmax estimator~\cite{jang2016categorical}. First, we draw a vector of independent and identically distributed (i.i.d.) uniform random variables and transform them into Gumbel$(0,1)$ noise using the inverse cumulative distribution function:
\begin{equation}
\mathbf{v}_{l,i} \sim \text{Uniform}(0,1)^{|\mathcal{A}|}, 
\qquad 
\mathbf{g}_{l,i} = -\log\!\Big(-\log(\mathbf{v}_{l,i}+\varepsilon)+\varepsilon\Big),
\end{equation}
where $\varepsilon$ is a small constant added for numerical stability.

The noise-perturbed logits are normalized via the softmax with temperature $\tau$:
\begin{equation}
y_{l,i,j} = 
\frac{\exp\!\left((\alpha_{l,i,j} + g_{l,i,j})/\tau\right)}
{\sum_{k=1}^{|\mathcal{A}|}\exp\!\left((\alpha_{l,i,k} + g_{l,i,k})/\tau\right)},
\qquad j = 1, \ldots, |\mathcal{A}|.
\end{equation}

To obtain a discrete selection while retaining differentiability during training, we adopt the straight-through (ST) Gumbel--Softmax estimator proposed by~\cite{jang2016categorical}:
\begin{equation}
\mathbf{h}_{l,i} = \operatorname{one\_hot}\!\big(\arg\max(\mathbf{y}_{l,i})\big) 
+ \big(\mathbf{y}_{l,i} - \operatorname{sg}(\mathbf{y}_{l,i})\big),
\end{equation}
where $\operatorname{sg}(\cdot)$ is the stop-gradient operator. In the forward pass, $\operatorname{sg}(\mathbf{y}_{l,i}) = \mathbf{y}_{l,i}$ in value, so $\mathbf{h}_{l,i}$ reduces to a hard one-hot vector. In the backward pass, the stop-gradient enforces $\tfrac{\partial \operatorname{sg}(\mathbf{y}_{l,i})}{\partial \mathbf{y}_{l,i}}=0$, so that
\begin{equation}
\frac{\partial \mathbf{h}_{l,i}}{\partial \mathbf{y}_{l,i}} = I.
\end{equation}
Here the one-hot term contributes no gradient, since $\arg\max$ is discrete and non-differentiable. As a result, the gradients propagate as if $\mathbf{h}_{l,i} = \mathbf{y}_{l,i}$, ensuring smooth optimization while retaining hard one-hot behavior in the forward pass.

In the forward pass, the output of neuron $i$ in layer $l$ is defined as the weighted sum of candidate activations

\begin{equation}
\text{activation } \eta_{l,i}(x_{l,i}) = \sum_{j=1}^{|\mathcal{A}|} h_{l,i,j} \cdot \sigma_j(x_{l,i})
\end{equation}

where $\mathbf{x}_{l,i}$ is the pre-activation input and $\sigma_j \in \mathcal{A}$ denotes the $j$-th activation function. Because $\mathbf{h}_{l,i}$ is a one-hot vector determined by the $\arg\max$ operation, the summation effectively collapses during the forward pass. This means the neuron behaves as if it has selected a single specialized function:
\begin{equation}
\text{activation } \eta_{l,i}(x_{l,i}) = \sigma_{\text{selected}}(x_{l,i})
\end{equation}

The Gumbel--Max trick ensures that the perturbed $\arg\max$ produces unbiased samples from the categorical distribution while the Gumbel--Softmax relaxation provides a differentiable approximation that enables gradient-based optimization of the logits $\boldsymbol{\alpha}_{l,i}$.
\begin{equation}
p_j = \frac{\exp(\alpha_{l,i,j})}{\sum_{k=1}^{|\mathcal{A}|}\exp(\alpha_{l,i,k})},
\qquad j = 1, \ldots, |\mathcal{A}|,
\end{equation}


\subsection{Phase 2: Mixed Network Construction}

After training for a predetermined number of epochs in the selective phase, we extract a single activation per neuron by taking the maximum-logit choice:
\begin{equation}
\hat{\sigma}_{l,i} \;=\; \arg\max_{j \in \{1,\ldots,|\mathcal{A}|\}} \alpha_{l,i,j},
\end{equation}
where $\alpha_{l,i,j}$ is the learned logit for neuron $i$ in layer $l$ associated with activation function $\sigma_j \in \mathcal{A}$, and $\hat{\sigma}_{l,i} \in \mathcal{A}$ denotes the final chosen activation for neuron $(l,i)$.

We then construct a fixed ``mixed'' network in which each neuron uses its selected activation. Let $n_l$ denote the number of neurons in layer $l$. For computational efficiency, neurons with the same chosen activation are grouped together:
\begin{equation}
\mathcal{G}_{l,\sigma} \;=\; \{\, i \in \{1,\ldots,n_l\} \;:\; \hat{\sigma}_{l,i}=\sigma \,\}, 
\qquad \sigma \in \mathcal{A}.
\end{equation}

In the forward pass of layer $l$, we first compute the pre-activation vector with a single affine transformation, using the trained weight matrix $\mathbf{W}_l \in \mathbb{R}^{n_l \times n_{l-1}}$ and bias vector $\mathbf{b}_l \in \mathbb{R}^{n_l}$:
\begin{equation}
\mathbf{u}_l \;=\; \mathbf{W}_l \mathbf{x}_{l-1} + \mathbf{b}_l \;\in\; \mathbb{R}^{n_l},
\end{equation}
where $\mathbf{x}_{l-1} \in \mathbb{R}^{n_{l-1}}$ is the input from the previous layer.

The layer output is computed efficiently by processing each activation function group separately:
\begin{equation}
(\mathbf{y}_l)_i \;=\; \sigma\!\big( (\mathbf{u}_l)_i \big), 
\qquad \forall i \in \mathcal{G}_{l,\sigma}, \; \forall \sigma \in \mathcal{A},
\end{equation}
where for each activation function $\sigma$, we apply $\sigma$ simultaneously to all neurons $i \in \mathcal{G}_{l,\sigma}$ that selected this activation function. This vectorized implementation processes entire neuron groups at once: for each $\sigma$, extract the pre-activations of neurons in $\mathcal{G}_{l,\sigma}$, apply $\sigma$ to this subvector, and assign the results back to their corresponding positions in $\mathbf{y}_l$.

After constructing the mixed network with fixed activation functions, we continue training the network for additional epochs. During this phase, the activation function selection remains fixed according to the choices made in Phase 1, while the network weights $\mathbf{W}_l$ and biases $\mathbf{b}_l$ continue to be optimized through standard backpropagation. This continued training allows the network parameters to adapt specifically to the selected activation configuration, often leading to improved performance as the weights can specialize for their assigned activation functions without the uncertainty introduced by the stochastic selection process of Phase 1.

\section{Experiments}
\label{sec:experiments}

\subsection{Data}

We conduct comprehensive experiments using the MNIST handwritten digit classification dataset, which consists of 70,000 grayscale images of digits 0--9, each with dimensions of 28$\times$28 pixels. 

To ensure robust evaluation and prevent overfitting, we employ stratified sampling to partition the original 60,000 training samples into two distinct subsets: 54,000 samples (90\%) for training and 6,000 samples (10\%) for validation. The official MNIST test set of 10,000 samples is used as the final test set for performance evaluation. All pixel values are normalized to the range [0, 1] by dividing by 255, and images are flattened to 784-dimensional vectors for input to fully connected networks. 

\subsection{Performance Analysis of SmartMixed Phases}

In this subsection, we aim to analyze whether SmartMixed is able to train a neural network effectively. We define and train 18 different feedforward neural network architectures based on SmartMixed (shown in \autoref{tab:architecture_rankings}). Our goal is neither to find nor to achieve the highest performance on the MNIST dataset, but rather to compare the performance of the same network architecture when trained using the SmartMixed approach versus fixed activation functions. We also investigate whether neurons in different layers and architectures exhibit systematic differences in their activation function selections. If such tendencies exist, we aim to identify and analyze them.

As explained in the previous section (i.e., \ref{sec:methodology}), the first phase of SmartMixed training focuses on enabling each neuron to discover its optimal activation function through differentiable selection. For analysis and illustration of the results, we use the architecture [784, 768, 512, 512, 256, 256, 128, 10] as an example. However, we observe that our findings are supported by the other architectures as well.

In the first phase of the training (i.e., the selective phase), we train the network for $\alpha$ epochs, where $\alpha$ represents the transition point between exploration and exploitation phases. In our experiments, we set $\alpha = 50$ epochs, though this parameter is adjustable based on the complexity of the architecture and convergence behavior. In future work, this parameter could be selected dynamically during training based on factors such as convergence in the loss function and the frequency of changes in activation function selections.

\begin{figure}[h]
\centering
\includegraphics[width=0.48\textwidth]{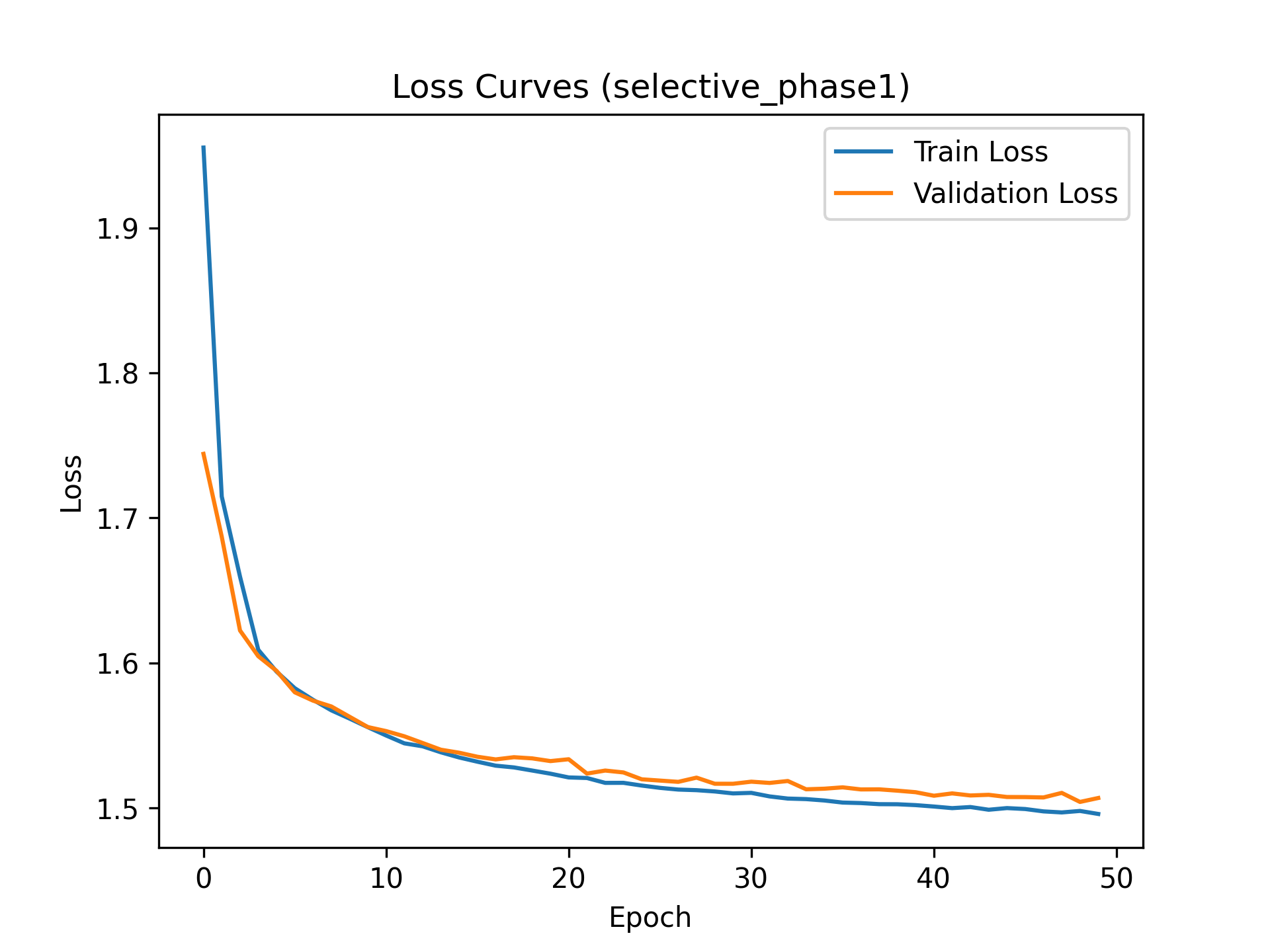}
\caption{Training and validation loss curves during Phase 1 (selective training).}
\label{fig:phase1_loss}
\end{figure}

\autoref{fig:phase1_loss} illustrates the training dynamics during the selective phase for the representative network architecture. The loss curves demonstrate healthy training behavior with a sharp initial decrease from approximately 2.0 to 1.5, followed by stabilization around epoch 30. The close alignment between training and validation curves throughout the process indicates effective learning without overfitting, confirming that the 50-epoch transition point provides sufficient time for activation preference stabilization.

The temperature parameter in the Gumbel-Softmax formulation plays a crucial role in controlling the sharpness of the learned activation selection. The hard discretization ensures that each neuron commits to a specific activation function for gradient computation based on its learned preferences. This design enables effective learning of activation preferences while maintaining gradient flow throughout the selective phase. The hyper-parameters used to train our neural network architectures are shown in  \autoref{tab:hyperparameters}.

\begin{figure}[h]
\centering
\includegraphics[width=0.48\textwidth]{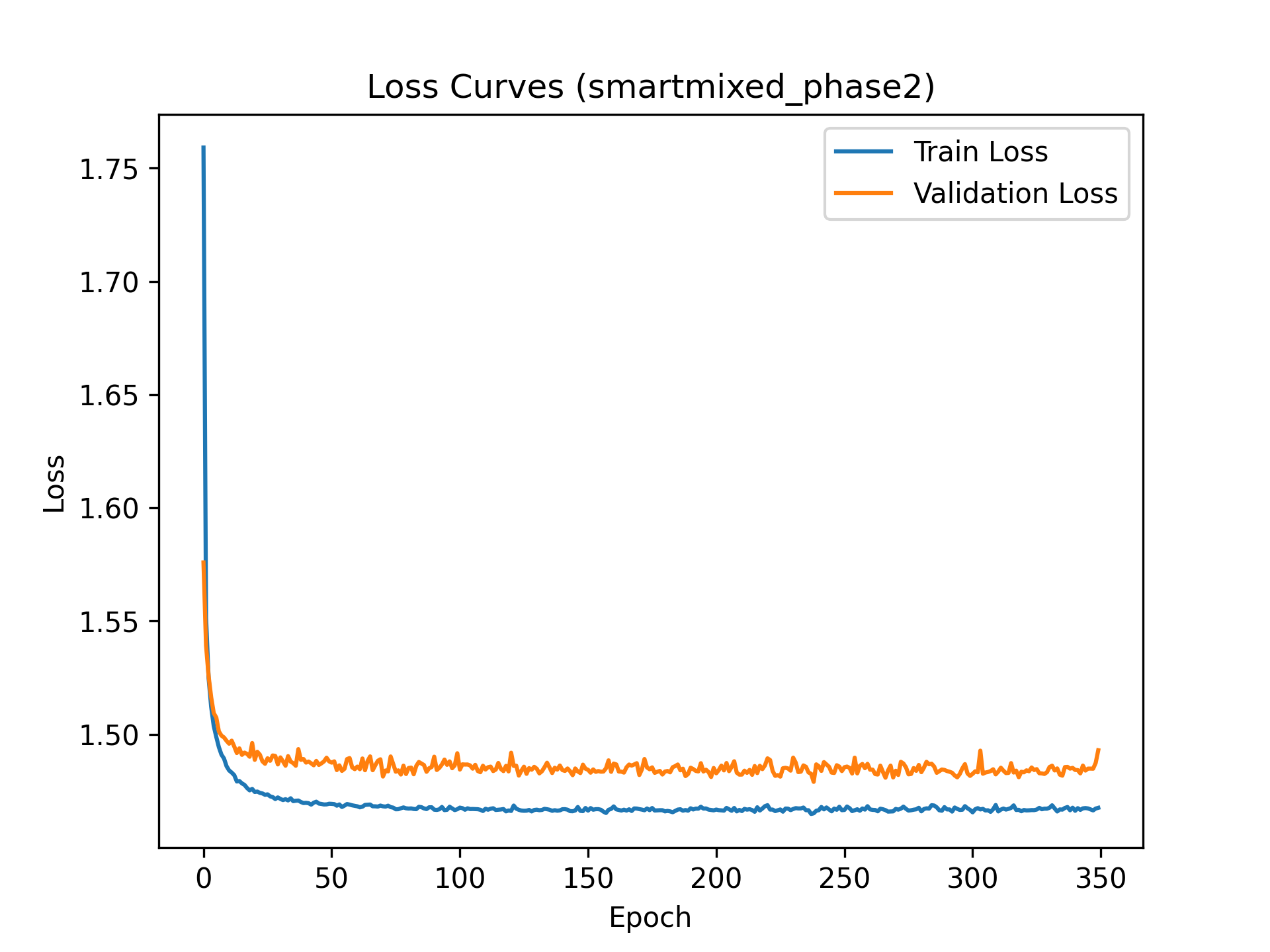}
\caption{Training and validation loss curves during Phase 2 (mixed network training).}
\label{fig:phase2_loss}
\end{figure}

Following Phase 1, we transition to Phase 2 by selecting the highest probability activation function for each neuron based on the learned logit preferences and reconstructing the network with these fixed activation choices. We then continue training this mixed network for an additional 350 epochs, primarily out of curiosity to observe the long-term training dynamics, though such an extended training period is not necessary for practical applications. \autoref{fig:phase2_loss} demonstrates that the Phase 2 training exhibits healthy behavior with stable convergence and close alignment between training and validation curves, confirming the effectiveness of the mixed network architecture. We analyse the training for the other architectures and observe similar healthy training behaviour. 

\begin{table*}[h]
\caption{Activation Function Performance Rankings Across 18 Network Architectures}
\label{tab:architecture_rankings}
\begin{tabular}{p{8cm}ccc}
\toprule
\textbf{Model Architecture} & \textbf{1st Rank} & \textbf{2nd Rank} & \textbf{3rd Rank} \\
\midrule
{[784, 512, 10]} & sigmoid & tanh & relu \\
{[784, 512, 256, 128, 10]} & smartmixed & leaky\_relu & relu \\
{[784, 512, 256, 128, 64, 10]} & smartmixed & leaky\_relu & tanh \\
{[784, 512, 256, 256, 128, 10]} & leaky\_relu & elu & smartmixed \\
{[784, 256, 128, 10]} & smartmixed & tanh & relu  \\
{[784, 768, 10]} & tanh & sigmoid & leaky\_relu \\
{[784, 768, 512, 10]} & relu & leaky\_relu & smartmixed \\
{[784, 768, 512, 512, 10]} & relu & smartmixed & leaky\_relu \\
{[784, 768, 512, 512, 256, 10]} & leaky\_relu & smartmixed & elu \\
{[784, 768, 512, 512, 256, 256, 10]} & leaky\_relu & relu & smartmixed \\
{[784, 768, 512, 512, 256, 256, 128, 10]} & smartmixed & relu & selu \\
{[784, 768, 512, 512, 256, 256, 128, 128, 10]} & leaky\_relu & smartmixed & relu \\
{[784, 768, 512, 512, 256, 256, 128, 128, 64, 10]} & relu & leaky\_relu & selu \\
{[784, 768, 512, 512, 256, 256, 128, 128, 64, 64, 10]} & relu & leaky\_relu & selu \\
{[784, 768, 512, 512, 256, 256, 128, 128, 64, 64, 32, 10]} & leaky\_relu & relu & smartmixed \\
{[784, 768, 512, 512, 256, 256, 128, 128, 64, 64, 32, 32, 10]} & relu & smartmixed & leaky\_relu \\
{[784, 768, 512, 512, 256, 256, 128, 128, 64, 64, 32, 32, 16, 10]} & leaky\_relu & relu & selu \\
{[784, 768, 512, 512, 256, 256, 128, 128, 64, 64, 32, 32, 16, 16, 10]} & smartmixed & leaky\_relu & relu \\
\bottomrule
\end{tabular}
\end{table*}

\begin{table}[htbp]
\centering
\caption{Training Hyperparameters Used in All Experiments}
\label{tab:hyperparameters}
\renewcommand{\arraystretch}{1.2} 
\begin{tabular}{ll}
\toprule
\textbf{Hyperparameter} & \textbf{Value} \\
\midrule
Optimizer               & Adam \\
Learning Rate           & $1 \times 10^{-4}$ \\
Batch Size              & 128 \\
Training Epochs         & 400 total (50 Phase 1, 350 Phase 2) \\
Gumbel-Softmax $\tau$   & 0.3 \\
Validation Split        & 10\% \\
\bottomrule
\end{tabular}
\end{table}

\begin{table}[h]
\caption{Performance Comparison Between Phase 1 and Phase 2}
\label{tab:phase_comparison}
\begin{tabular}{lcc}
\toprule
\textbf{Training Phase} & \textbf{Validation Accuracy} & \textbf{Test Accuracy} \\
\midrule
Phase 1 (Selective) & 95.67\% & 95.73\% \\
Phase 2 (Mixed) & 98.22\% & 98.03\% \\
\bottomrule
\end{tabular}
\end{table}

We can observe in \autoref{tab:phase_comparison}, phase 2 achived approximately 2--3\% improvement in accuracy over Phase 1. This improvement is particularly significant because it occurs after fixing the activation functions, indicating that the learned activation choices provide a better foundation for continued training. The stabilization of activation functions in Phase 2 is beneficial as any changes in activation function selection would require network weight readjustment due to different output ranges. By fixing the activation choices based on learned preferences, the network can focus entirely on optimizing the weights for the selected activation functions, leading to more stable and effective learning.

\subsection{Analysis of Activation Function Selection}

\begin{figure*}[h]
\centering
\includegraphics[width=0.8\textwidth]{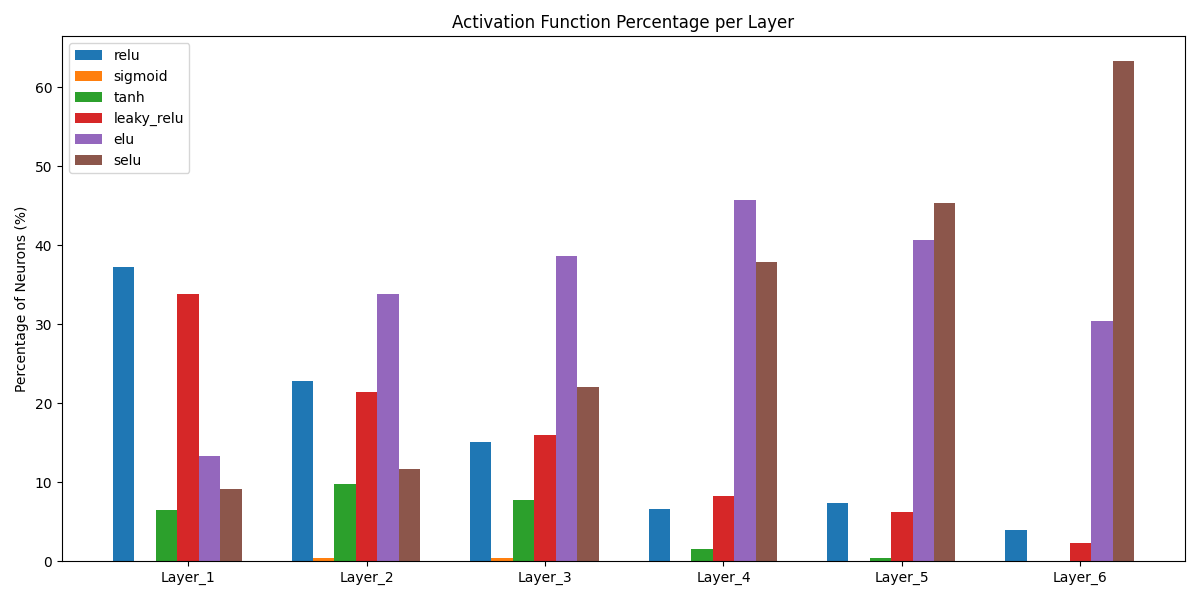}
\caption{Final activation function distribution across layers after Phase 1 training. Early layers favor ReLU and Leaky\_ReLU, while deeper layers increasingly prefer ELU and SELU activation functions.}
\label{fig:activation_bar_chart}
\end{figure*}

In this analysis, we examine whether neurons in different layers exhibit distinct behaviors when selecting their activation functions. As an example, we use the same representative network architecture from the previous section [784, 768, 512, 512, 256, 256, 128, 10], focusing exclusively on the six hidden layers and excluding the input and output layers from activation function selection. In the following analysis, ``Layer 1'' refers to the first hidden layer (768 neurons), ``Layer 2'' to the second hidden layer (512 neurons), and so forth, with ``Layer 6'' representing the final hidden layer (128 neurons) before the output layer. \autoref{fig:activation_bar_chart} presents the final activation function distribution across all six hidden layers after 50 epochs of Phase 1 training, revealing distinct layer-wise preferences.

\begin{figure}[!t]
\centering
\includegraphics[width=0.48\textwidth]{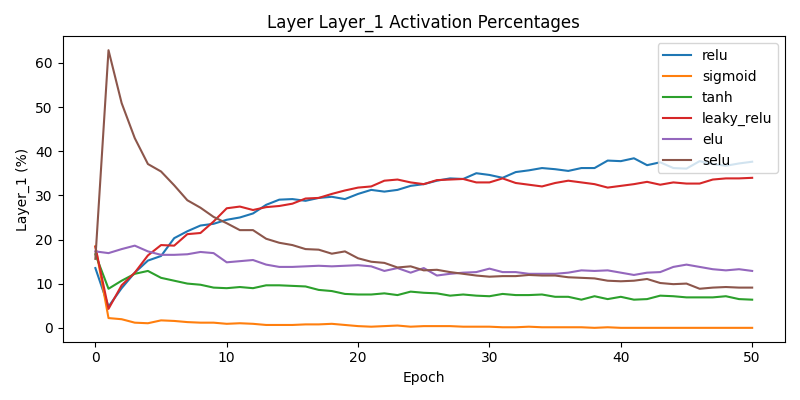}
\caption{Activation function distribution of neurons in layer 1 during each epoch of Phase 1 training. As the number of epochs increases, the number of neurons selecting ReLU or Leaky\_ReLU grows, while the selection of other activation functions decreases. }
\label{fig:layer1_over_epochs}
\end{figure}

\begin{figure}[!t]
\centering
\includegraphics[width=0.48\textwidth]{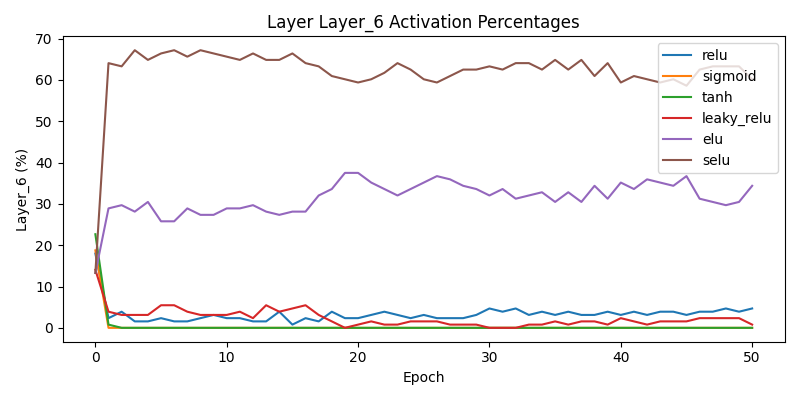}
\caption{Activation function distribution of neurons in layer 6 during each epoch of Phase 1 training.}
\label{fig:layer6_over_epochs}
\end{figure}

The results reveal clear patterns in activation function preferences across different network layers. Layer 1 neurons strongly prefer ReLU ($\sim$37\%) and Leaky\_ReLU ($\sim$34\%), accounting for over 70\% of selections. Notably, Sigmoid is consistently avoided across all layers, likely due to its well-documented gradient vanishing problems that impede effective learning in deep networks. 

\autoref{fig:layer1_over_epochs} shows how neurons’ activation function selections change across epochs. As the number of epochs increases, the number of neurons selecting either ReLU or Leaky\_ReLU as their activation function grows, while the selection frequency of other activation functions decreases.

As network depth increases, preferences shift dramatically toward ELU and SELU, with layer 6 showing the most pronounced preference for SELU ($\sim$64\%) and ELU ($\sim$31\%). As shown in \autoref{fig:layer6_over_epochs}, even after a few epochs, the majority of neurons have already chosen SELU and ELU as their activation functions.

This analysis demonstrates that neurons in different layers have distinct preferences when selecting activation functions, and SmartMixed allows them to follow these preferences while the network is trained efficiently. The neurons exhibit a clear pattern where early layers favor ReLU and Leaky\_ReLU, whereas deeper layers prefer ELU and SELU. To validate the generalizability of these findings, we also analyzed the activation preferences of neurons in other deep neural network architectures listed in \autoref{tab:architecture_rankings}.

The results of these experiments consistently reveal similar layer-wise activation preferences, confirming the robustness of our observations. Specifically, the distribution of selected activation functions supports the layer-wise trends identified in the previous subsection in a way that deeper layers consistently favor ELU and SELU over alternatives such as ReLU, Sigmoid, and Tanh. This consistency across diverse architectural configurations demonstrates both the reliability of the learned activation preferences and the generalizability of our findings. To the best of our knowledge, such consistent layer-wise specialization patterns have not been explicitly reported in prior literature. A comprehensive analysis of performance and architectural comparisons is presented in the following section.

\subsection{Evaluating SmartMixed across Diverse Network Topologies}

It is worth mentioning again that the objective of our architectural evaluation is not to find the best architecture that achieves the highest accuracy on the MNIST dataset, but rather to investigate how does SmartMixed's adaptive approach compare against traditional fixed activation function strategies across diverse architectural configurations.

To answer this question, as we mentioned earlier, we conduct experiments on 18 distinct architectures, as shown in \autoref{tab:architecture_rankings}. For each architecture, we train seven separate models: six using fixed activation functions (ReLU, Sigmoid, Tanh, Leaky ReLU, ELU, SELU) and one using our proposed SmartMixed approach. The models are then ranked based on their test accuracy.

The results shown in \autoref{tab:architecture_rankings} reveal that SmartMixed demonstrates competitive performance, consistently appearing in the top three rankings across the majority of tested architectures. Notably, the performance differences among most activation functions are relatively small. However, Sigmoid consistently exhibits significantly lower performance, particularly in deeper architectures.

\begin{figure}[!t]
\centering
\includegraphics[width=0.48\textwidth]{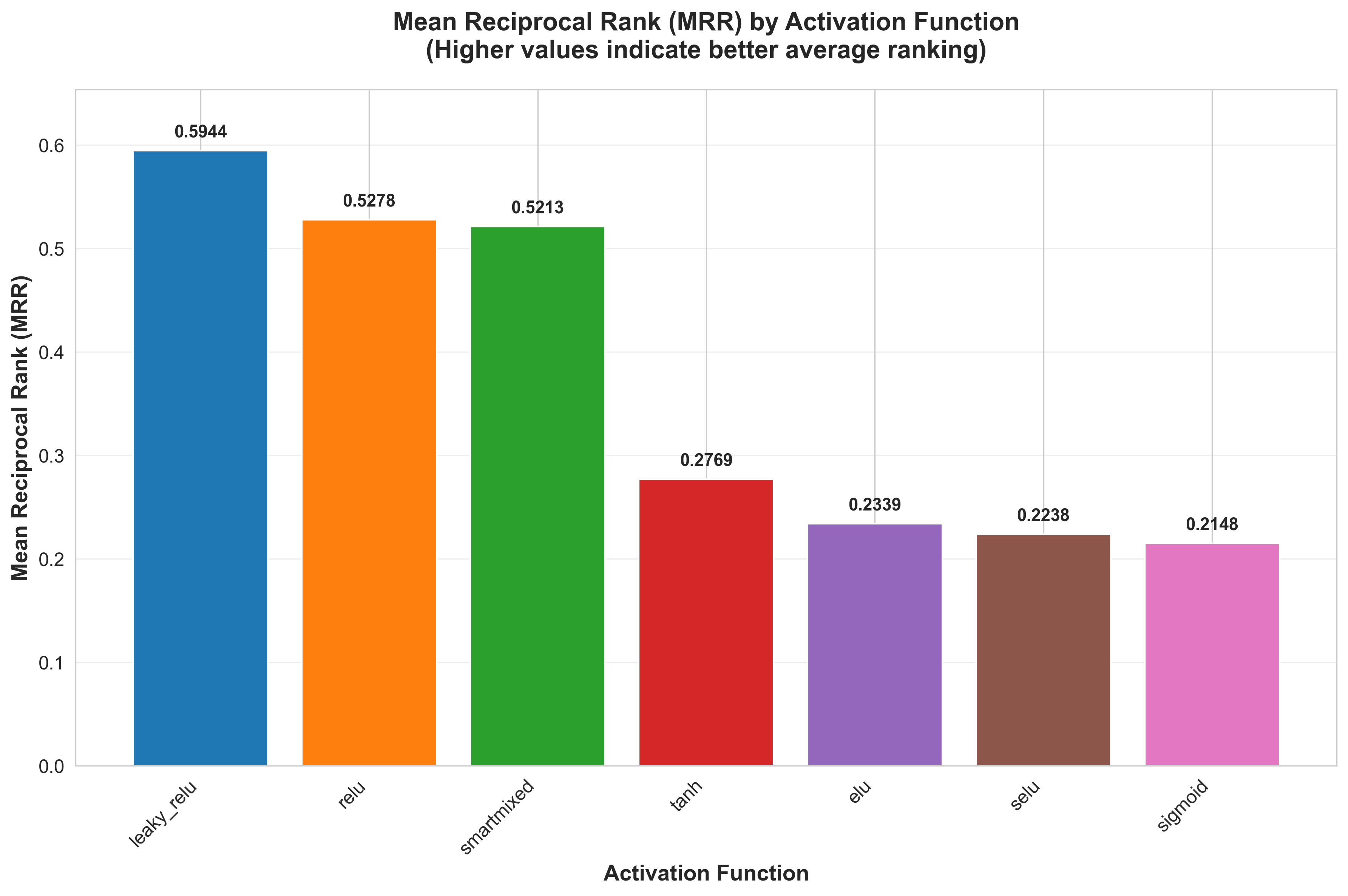}
\caption{Mean Reciprocal Rank analysis across all architectures. Leaky\_ReLU demonstrates the highest MRR, followed closely by ReLU and SmartMixed. SmartMixed shows consistent performance by ranking in top positions across diverse architectures.}
\label{fig:mrr_chart}
\end{figure}

\autoref{fig:mrr_chart} presents the Mean Reciprocal Rank (MRR) analysis, which provides a comprehensive measure of the average performance of each activation function across all architectures. The MRR is calculated as:

\begin{equation}
\text{MRR} = \frac{1}{n} \sum_{i=1}^{n} \frac{1}{\text{rank}_i}
\end{equation}

where $n$ is the total number of architectures and $\text{rank}_i$ is the rank of the activation function for architecture $i$. Higher values indicate better overall performance. Based on the ranking patterns, Leaky\_ReLU and ReLU show the strongest consistent performance, while SmartMixed demonstrates competitive results by consistently ranking in the top positions even when not achieving first place. The analysis shows that while no single activation function dominates universally, SmartMixed provides a robust adaptive solution that performs well across diverse architectural configurations.

\begin{figure*}[h]
\centering
\includegraphics[width=0.8\textwidth]{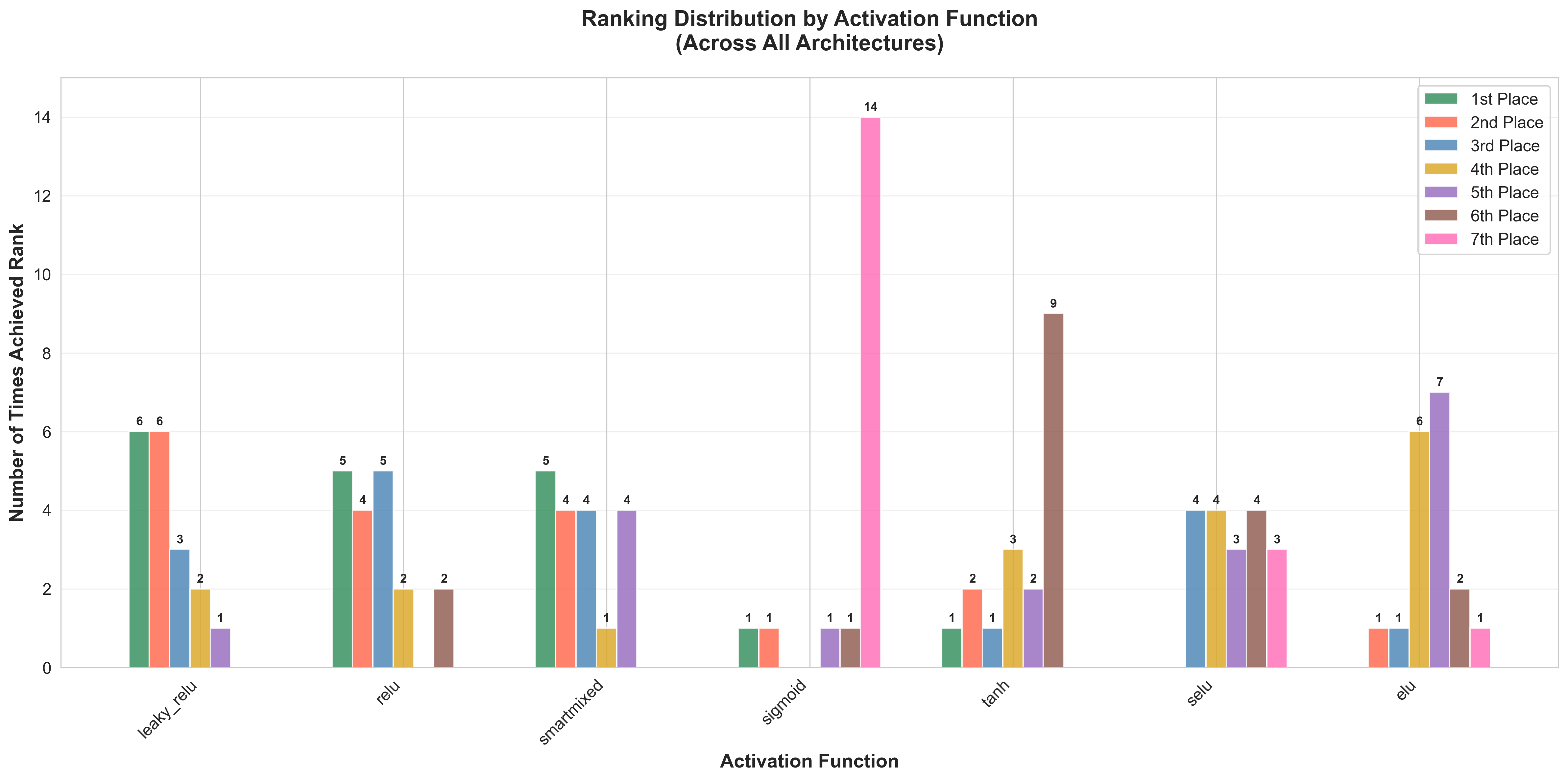}
\caption{Ranking distribution showing the frequency of each rank position (1st through 7th) achieved by each activation function across different architectures. SmartMixed demonstrates consistent performance with frequent appearances in top positions.}
\label{fig:ranking_distribution}
\end{figure*}

\autoref{fig:ranking_distribution} complements the MRR analysis by showing the frequency distribution of rankings achieved by each activation function across all architectures. The results reveal a clear performance hierarchy: Leaky\_ReLU, ReLU, and SmartMixed emerge as the top-performing approaches, frequently competing for first-place rankings and consistently appearing within the top three positions. In contrast, traditional activation functions such as Sigmoid and Tanh consistently underperform, predominantly occupying lower-ranked positions across most architectures. This distribution pattern further validates SmartMixed's effectiveness as an adaptive approach capable of competing with well-established fixed activation functions.

\begin{figure}[!t]
\centering
\includegraphics[width=0.48\textwidth]{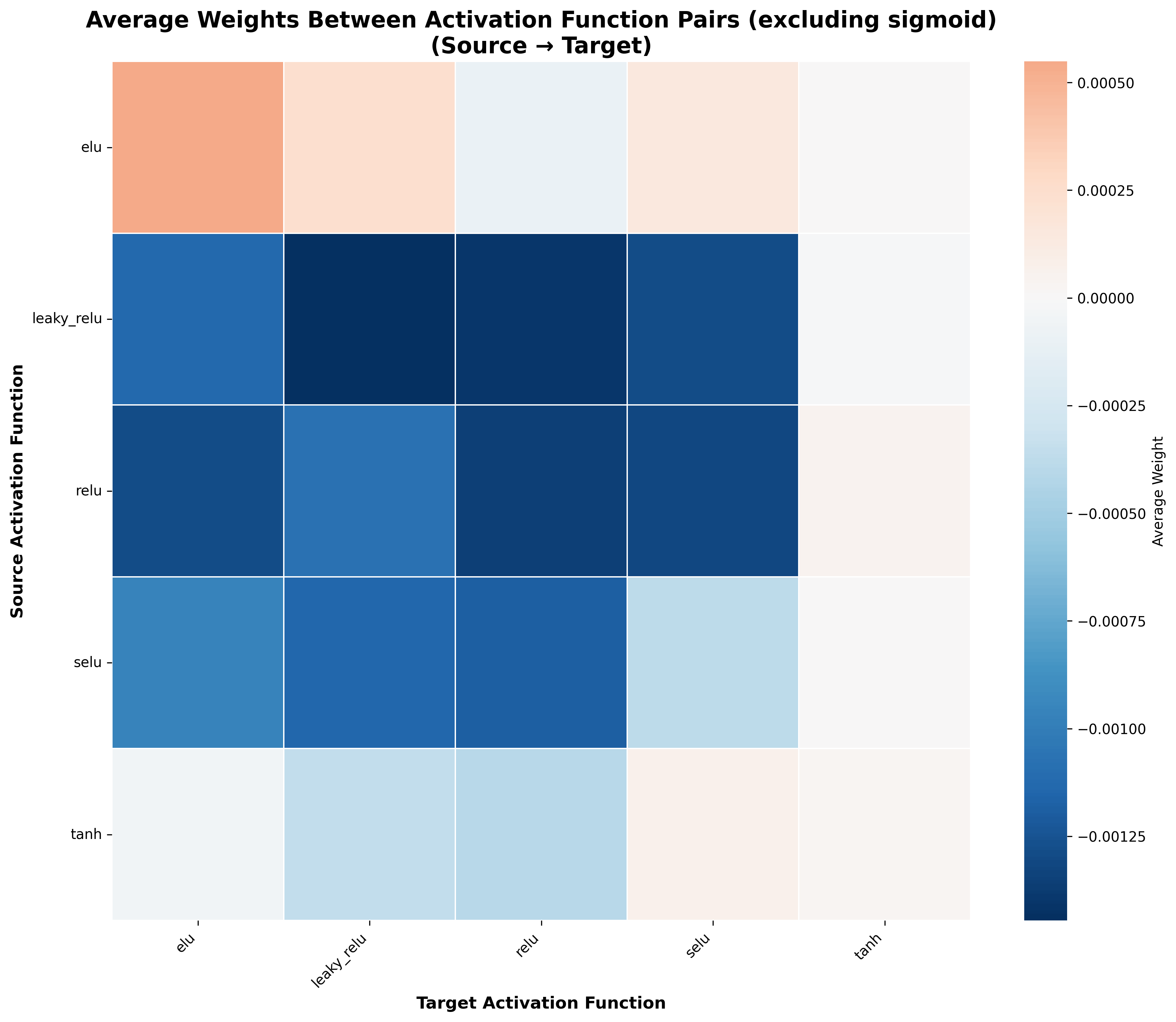}
\caption{Heatmap of average connection weights between neurons grouped by activation function types across different architectures. Sigmoid is excluded due to infrequent selection. The visualization shows weight patterns between source neurons (vertical axis) and target neurons (horizontal axis), revealing connectivity preferences between different activation function types.}
\label{fig:activation_weights_heatmap}
\end{figure}

To further understand the internal workings of SmartMixed networks, we analyzed the weight connectivity patterns between neurons with different activation functions across all architectures. \autoref{fig:activation_weights_heatmap} presents a heatmap visualization of the average connection weights between neurons grouped by their activation function types, with Sigmoid excluded due to its rare selection frequency across the networks.

The heatmap reveals distinct patterns in inter-activation connectivity, showing how neurons with different activation functions connect through learned weight matrices. The source activation function (vertical axis) represents the sending neuron while the target activation function (horizontal axis) represents the receiving neuron, with color intensity indicating average connection weights. A notable observation is that neurons receiving inputs from ELU neurons exhibit consistently more positive average weights compared to other activation function pairs. To the best of our knowledge, this finding is novel and has not been seen in other publications.

\section{Conclusion}
\label{sec:conclusion}

We introduced SmartMixed, a novel two-phase training strategy that enables practical per-neuron activation function learning in neural networks. Our approach successfully bridges the gap between adaptive and fixed activation approaches, providing the benefits of both while mitigating their respective limitations. The process encompasses two phases: the selective phase enables thorough exploration of activation function choices using differentiable Gumbel-Softmax sampling, while the mixed phase exploits learned preferences through efficient vectorized operations. Experimental results on MNIST demonstrate the effectiveness of SmartMixed in training feedforward neural networks, by appearing among the top three performing approaches across 18 diverse network architectures, validating its robustness and adaptability.

Most significantly, our analysis discovered intrinsic activation function preferences that emerge at different network depths in a way that neurons in different layers exhibit distinct and consistent preferences: early layers strongly favor ReLU and Leaky\_ReLU activation functions while deeper layers progressively shift toward ELU and SELU functions. This layer-wise specialization pattern remained consistent across diverse architectural configurations. These findings provide novel insights into the functional diversity within neural architectures and demonstrate that individual neurons can intelligently adapt their activation functions based on their positional role in the network.

Moreover, SmartMixed opens new avenues for neural architecture optimization by demonstrating that sophisticated architectural choices can be learned during training rather than being selected as hyperparameter, without sacrificing deployment efficiency. The principles underlying our approach are broadly applicable and could inspire similar staged optimization strategies for other architectural components.

Another interesting direction for future work would be to evaluate SmartMixed on more complex and diverse datasets, which could provide insights into the generalizability of the observed activation function preferences across different data modalities and complexity levels. Additionally, extending our approach to other types of neural networks, especially state-of-the-art architectures, would allow us to investigate whether the patterns observed in feedforward networks hold across various neural network types.



\bibliographystyle{ACM-Reference-Format}
\bibliography{references}

\end{document}